\RequirePackage[T1]{fontenc}
\documentclass[letterpaper,10pt,conference]{ieeeconf}
\IEEEoverridecommandlockouts
\usepackage{times}
\usepackage{amsmath,amssymb}
\usepackage{graphicx}
\usepackage{float}
\usepackage{booktabs}
\usepackage{cite}
\usepackage{xcolor}
\usepackage{url}
\newcommand{\lifd}{\textsc{LIFD}}
\newcommand{\sg}{\operatorname{sg}}
\newcommand{\attn}{\operatorname{Attn}}
\newcommand{\film}{\operatorname{FiLM}}

\title{LIFD: Anchored Diffusion for\\
3D-Aware Scene Memory in Robotic Manipulation}
\author{%
\authorblockN{Wenbo Li$^{1}$, Yiteng Chen$^{1}$, Wenhao Li$^{1}$, and Qingyao Wu$^{1,\dagger}$}
\authorblockA{$^{1}$School of Software Engineering, South China University of Technology}
\thanks{$^{\dagger}$Corresponding author: Qingyao Wu.}}
\begin{document}
\bstctlcite{IEEEauthorlistcontrol}
\maketitle
\thispagestyle{empty}
\pagestyle{empty}
\begin{abstract}
During manipulation, robot and scene motion can move previously observed regions outside the camera's field of view.
Geometry-aware RGB features encode visible structure, while control under partial observability requires scene memory that integrates observation history and grounds inferred content in current evidence.
We introduce \lifd{} (Look, Imagine, Focus, and Do), a framework for persistent, 3D-aware scene memory.
LIFD learns scene tokens through multi-view agreement, then completes them from a single RGB view and recurrent memory using rectified flow.
Anchor-Guided Cross-Attention anchors generation to current geometry-aware features, and compact slot features condition a visuomotor policy.
Multi-view and geometric supervision are used during representation learning; deployment requires one RGB camera, proprioception, and a task instruction.
LIFD (Staged) reaches 91.6\% average success on LIBERO and 79.8\% on MetaWorld, improving LIBERO average success by 11.1 percentage points over Joint training.
After policy-head adaptation with ten demonstrations per family, LIFD achieves 56.0\% mean success across four UR5e task families, compared with 40.5\% for OpenVLA-7B.
\end{abstract}

\section{Introduction}
\label{sec:intro}
Robotic manipulation in cluttered, partially observable environments requires a persistent scene representation.
An object can remain relevant to an action after it leaves the current field of view, while manipulation itself changes what the camera observes.
The scene representation must integrate newly visible structure with observation history.

Point-cloud and voxel policies encode spatial structure for manipulation~\cite{shridhar2023perceiver,ze20243d,jia2025lift3d}.
Their geometric inputs require depth sensing or reconstruction.
Geometry-aware image encoders offer a route to 3D priors from RGB~\cite{wang2025vggt}, but the current view still leaves parts of the scene unobserved.
Recent memory-based policies retain observation history and spatial context~\cite{fang2025sam2act,shi2025memoryvla,li2026bridgevlapp}.

Generative models provide learned priors for inferring unobserved content~\cite{ha2018world,seo2023masked}.
A plausible completion can still conflict with visible structure, so inferred scene content must remain grounded in current geometric evidence and observation history.

We introduce \lifd{} (Look, Imagine, Focus, and Do), a framework for persistent, 3D-aware scene memory.
Multi-view self-distillation encourages scene-token agreement across views; anchored rectified flow completes these tokens from current geometry-aware RGB features and recurrent memory.
\emph{Look} and \emph{Imagine} construct this representation; \emph{Focus} condenses it into slot features, and \emph{Do} predicts actions with a lightweight visuomotor policy.
Multi-view and geometric supervision shape the representation during training, while deployment uses a single camera stream.

We evaluate LIFD on LIBERO, MetaWorld, RoboTwin 2.0, and four UR5e task families.
Ablations and observation interventions assess how generative completion, visible anchoring, and observation history contribute to control; latency measurements characterize the cost of iterative inference.

\begin{figure*}[t]
\centering
\includegraphics[width=0.90\textwidth]{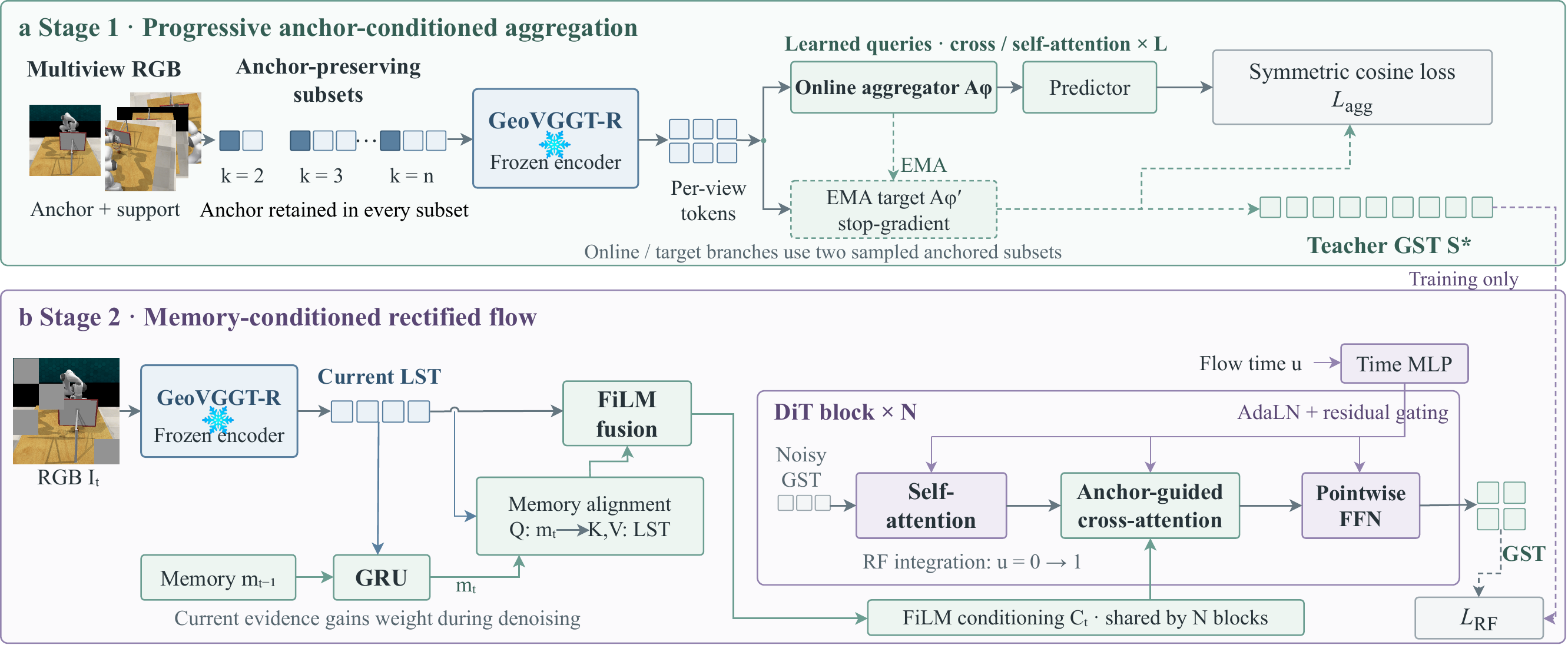}
\caption{\textbf{Staged representation learning.} (a) Anchor-conditioned aggregation and self-distillation learn multi-view scene tokens. (b) Rectified flow completes them conditioned on current-view features and recurrent memory, progressing from noise at $u=0$ to teacher GST at $u=1$.}
\label{fig:lifd_overview}
\end{figure*}

\section{Related Work}
\label{sec:related}
\subsection{Visual Representations for Manipulation}
Visual pretraining supplies transferable features for visuomotor learning~\cite{nair2022r3m,majumdar2023we}.
Point-cloud and voxel policies encode spatial structure for manipulation~\cite{shridhar2023perceiver,ze20243d,jia2025lift3d}.
Geometry-aware image encoders such as VGGT predict camera parameters, depth, point maps, and tracks from image collections~\cite{wang2025vggt}.
LIFD combines these geometric priors with multi-view supervision to learn a scene-token representation for deployment with single-view RGB observations.

\subsection{Memory for Robotic Manipulation}
Memory-based policies encode observation history to address partial observability.
TraceVLA overlays tracked point trajectories on current images~\cite{zheng2024tracevla}, while MemoryVLA stores perceptual and semantic features for retrieval and fusion~\cite{shi2025memoryvla}.
SAM2Act+ extends a 3D manipulation policy with memories of visual features and predicted action heatmaps~\cite{fang2025sam2act}.
BridgeVLA++ combines temporal context with spatial memory constructed from initial point clouds~\cite{li2026bridgevlapp}.
In LIFD, recurrent memory and current RGB features jointly condition generative completion toward a multi-view scene-token target.

\subsection{Generative Representations and Control}
World models learn environment representations and dynamics for planning and control~\cite{ha2018world,hafner2019learning,seo2023masked}.
Diffusion-based policies generate actions or trajectories~\cite{janner2022planning,chi2025diffusion}, and WorldVLA jointly models action generation and future images~\cite{cen2025worldvla}.
LIFD completes the current scene representation for downstream control.
Its multi-view targets build on set-latent aggregation~\cite{sajjadi2022scene,jaegle2021perceiver}, while conditional completion in token space is learned with a rectified-flow objective~\cite{liu2022flow}.

\newpage
\section{Method}
\label{sec:method}
\begin{figure*}[t]
\centering
\includegraphics[width=\textwidth]{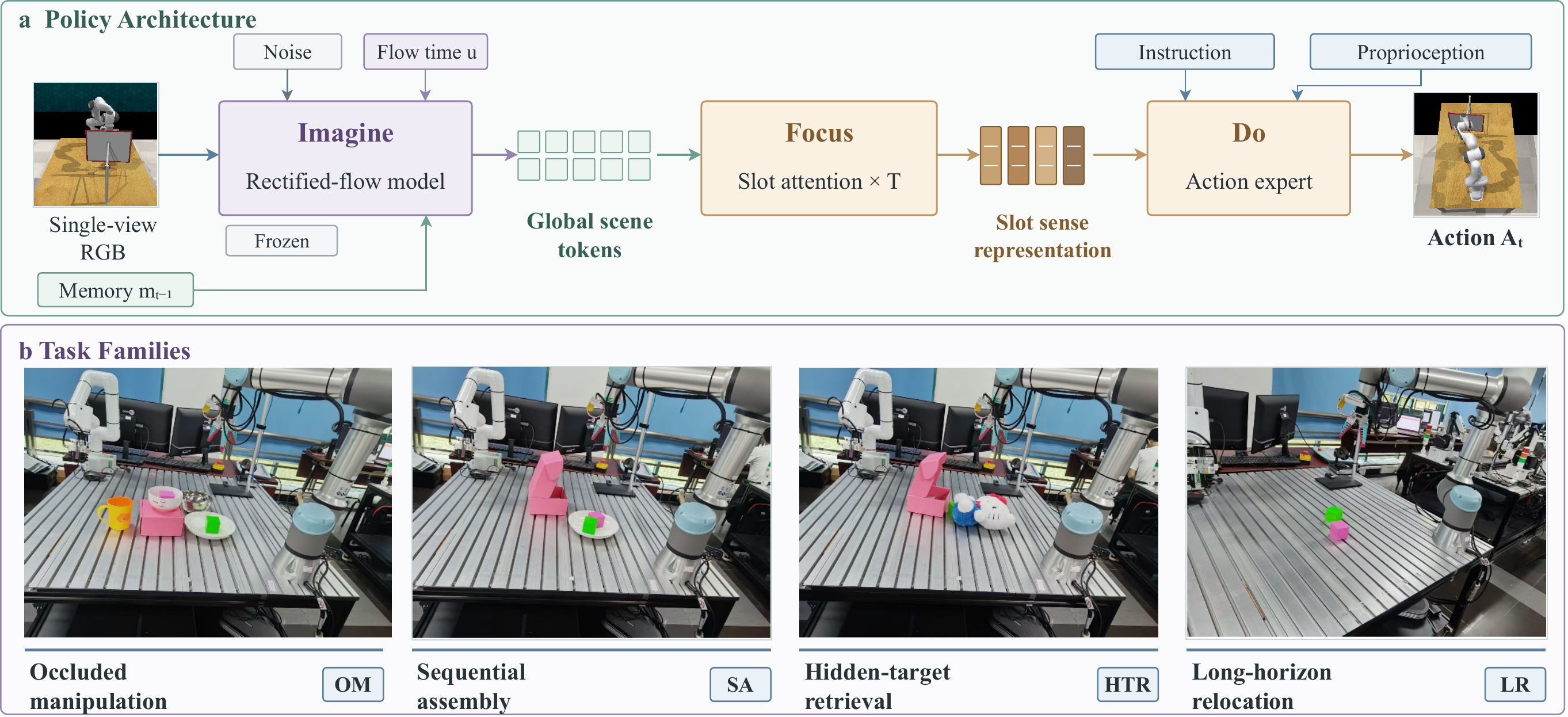}
\caption{\textbf{Policy and physical tasks.} (a) Imagine completes scene tokens, Focus groups them into object-centric slots, and Do predicts actions. (b) UR5e task families: Occluded Manipulation (OM), Sequential Assembly (SA), Hidden-Target Retrieval (HTR), and Long-Horizon Relocation (LR).}
\label{fig:stage3_pipeline}
\label{fig:tasks}
\end{figure*}

LIFD learns a scene representation from multi-view supervision and completes it at deployment from a single RGB stream and observation history.
At environment step $t$, a geometry-aware encoder $E$ maps the image $\mathbf I_t$ to Local Sense Tokens (LST), $\mathbf T_t\in\mathbb R^{N\times d}$.
Recurrent memory and current LST condition the generation of Global Scene Tokens (GST), $\widehat{\mathbf S}_t\in\mathbb R^{M\times d}$, which supply the policy with a scene-level representation.
Here $N$ and $M$ are token counts and $d$ is the feature dimension.
GST encode geometry in latent features rather than explicit metric coordinates.
Proprioception $\mathbf P_t$ and instruction embedding $\mathbf e_L$ enter the policy head.
Fig.~\ref{fig:lifd_overview} summarizes representation learning; Fig.~\ref{fig:stage3_pipeline} shows the deployed information flow.

\subsection{Learning a Geometry-Aware Scene Representation}
\label{sec:teacher}
\textbf{From local geometry to scene-level targets.}
GeoVGGT-R is obtained by LoRA-adapting a pretrained VGGT backbone~\cite{wang2025vggt} with camera, depth, point-map, and tracking supervision, and is frozen in subsequent stages.
Geometric adaptation encodes spatial structure in RGB features; multi-view aggregation combines evidence beyond an individual view.
The supervision heads and loss definitions are given in Appendix~\ref{app:implementation}.

For views $\mathcal I=\{I_a,I_2,\ldots,I_n\}$, every training subset contains the same anchor image $I_a$; the number of support views increases progressively from one to $n-1$.
For an anchor-containing subset $C\subseteq\mathcal I$, $A_\phi$ aggregates encoded view features with learned queries:
\begin{equation}
\mathbf S(C)=A_\phi\big(\{E(I):I\in C\}\big),\qquad I_a\in C
\label{eq:aggregation}
\end{equation}
Cross-attention aggregates view features, and self-attention mixes query features.
A fixed query count keeps the scene-token target size constant as support views increase.
The shared anchor gives different subsets a common visual reference.
To encourage agreement despite changing support views, two subsets $C_A,C_B$ are aligned by a symmetric self-distillation objective~\cite{grill2020bootstrap,chen2021exploring}:
\begin{equation}
\begin{gathered}
\mathbf p_A=h_\phi(\mathbf S(C_A)),\quad
\mathbf z_B=\sg\!\left[A_{\phi'}(E(C_B))\right],\\
\mathcal L_{\rm agg}=D(\mathbf p_A,\mathbf z_B)+D(\mathbf p_B,\mathbf z_A)
\end{gathered}
\label{eq:distillation}
\end{equation}
The terms $\mathbf p_B,\mathbf z_A$ swap $A$ and $B$.
Here $E(C)$ collects the encoded views, $h_\phi$ is the online predictor, $\sg$ stops gradients, and $D$ is cosine distance between corresponding scene-token arrays.
The target parameters follow an exponential moving average (EMA), $\phi'\leftarrow\beta\phi'+(1-\beta)\phi$~\cite{grill2020bootstrap}.
The trained EMA aggregator supplies multi-view scene-token targets $\mathbf S_t^\star$ for completion; its extra views are used only during learning and offline diagnostics.

\subsection{Anchored Completion with Temporal Memory}
\label{sec:completion}
\textbf{Retaining history and incorporating current evidence.}
A gated recurrent unit (GRU)~\cite{chung2014empirical} retains observation history for scene-token completion; memory-to-LST attention aligns its state with current-view features before fusion:
\begin{equation}
\begin{gathered}
\mathbf m_t=\operatorname{GRU}(\mathbf T_t,\mathbf m_{t-1}),\\
\overline{\mathbf m}_t=\attn(\mathbf m_t,\mathbf T_t,\mathbf T_t)
\end{gathered}
\label{eq:memory}
\end{equation}
Memory updates once per environment observation, indexed by $t$; $u\in[0,1]$ denotes generative flow progress.

\textbf{Visible-feature anchoring.}
FiLM modulation~\cite{perez2018film} fuses current-view features and recurrent memory after mapping the two arrays to compatible dimensions:
\begin{gather}
\mathbf C_t(u)=\alpha(u)\film_{\rm vis}(\mathbf T_t)
 +(1-\alpha(u))\film_{\rm mem}(\overline{\mathbf m}_t),\notag\\
\alpha(u)=\alpha_{\min}+u(\alpha_{\max}-\alpha_{\min})
\label{eq:fusion}
\end{gather}
With $0\leq\alpha_{\min}\leq\alpha_{\max}\leq1$, the weight on current visual evidence is nondecreasing along the flow.
Anchor-Guided Cross-Attention (AGCA) injects this conditioning into every block of a DiT-style transformer~\cite{peebles2023scalable}.
For evolving hidden tokens $\mathbf H$,
\begin{equation}
\mathbf H\leftarrow\mathbf H+
 \operatorname{MHA}\!\left(\operatorname{LN}(\mathbf H),
 W_K\mathbf C_t(u),W_V\mathbf C_t(u)\right)
\label{eq:agca}
\end{equation}
Here MHA and LN denote multi-head attention and layer normalization.
Self-attention mixes scene-token features; cross-attention injects $\mathbf C_t(u)$.

\textbf{Learning to complete scene tokens.}
The network learns a rectified flow (RF)~\cite{liu2022flow} from Gaussian noise $\boldsymbol\epsilon\sim\mathcal N(0,\mathbf I)$ to the teacher representation:
\begin{equation}
\begin{gathered}
\mathbf X_t(u)=(1-u)\boldsymbol\epsilon+u\mathbf S_t^\star,\\
\mathcal L_{\rm RF}=\mathbb E\left[
\left\|v_\theta(\mathbf X_t(u),u,\mathbf C_t(u))
 -(\mathbf S_t^\star-\boldsymbol\epsilon)\right\|_F^2\right]
\end{gathered}
\label{eq:flow}
\end{equation}
The expectation covers training observations, noise, and sampled flow progress.
At inference, integrating the learned vector field from $u=0$ to $u=1$, with classifier-free guidance (CFG)~\cite{ho2022classifier}, produces $\widehat{\mathbf S}_t$.
The number of integration steps $J$ controls inference cost; within these steps, memory stays fixed while its weighting against visible features changes with $u$.
Appendix~\ref{app:implementation} gives the attention, modulation, and guidance details.

\begin{table*}[t]
\caption{Main results on LIBERO, MetaWorld, and RoboTwin 2.0 under the corresponding benchmark settings. LIFD suite-level LIBERO and MetaWorld entries report mean $\pm$ SE; LIBERO Avg averages the four suite means. Other methods report means. SCI ($\downarrow$) is evaluated on LIBERO-Spatial for LIFD variants. $\dagger$: 3D-policy architectures adapted to the single-view RGB evaluation interface in LIBERO and the physical evaluation; RoboTwin entries retain published settings. --: unreported or inapplicable.}
\label{tab:main_results}
\centering
\small
\setlength{\tabcolsep}{1.8pt}
\begin{tabular*}{\textwidth}{@{\extracolsep{\fill}}lccccccccc@{}}
\toprule
&\multicolumn{5}{c}{LIBERO SR (\%)$\uparrow$}&MetaWorld&SCI$\downarrow$&\multicolumn{2}{c}{RoboTwin 2.0 SR (\%)$\uparrow$}\\
\cmidrule(lr){2-6}\cmidrule(lr){9-10}
Method&Spatial&Object&Goal&Long&Avg&Avg. SR (\%)$\uparrow$&&\shortstack{Hard\\(Clean$\rightarrow$Random)}&\shortstack{Easy\\(Clean$\rightarrow$Clean)}\\
\midrule
BC-R3M&$54.0$&$51.0$&$43.0$&$41.0$&$47.3$&--&--&--&--\\
VC-1&$58.0$&$56.0$&$45.0$&$43.0$&$50.5$&--&--&--&--\\
PerAct$^\dagger$&$61.5$&$57.8$&$49.3$&$47.0$&$53.9$&--&--&--&--\\
ACT~\cite{zhao2023act}&82.0&78.8&66.1&44.0&67.7&--&--&1.7&29.7\\
DP3$^\dagger$&$63.2$&$59.5$&$51.0$&$49.1$&$55.7$&--&--&$5.0$&$55.2$\\
Lift3D$^\dagger$&$64.0$&$60.2$&$51.8$&$50.0$&$56.5$&--&--&--&--\\
Diffusion Policy&$66.0$&$62.0$&$54.0$&$52.0$&$58.5$&$10.5$&--&$0.6$&$28.0$\\
OpenVLA-7B&$77.5$&$73.5$&$64.1$&$60.3$&$68.9$&--&--&--&--\\
OpenVLA-OFT~\cite{kim2025oft}&$97.6$&$98.4$&$97.9$&$94.5$&$97.1$&--&--&--&--\\
RDT-1B~\cite{liu2024rdt}&60.2&77.8&68.2&29.0&58.8&--&--&13.7&34.5\\
$\pi_0$&$79.2$&$75.0$&$65.5$&$61.7$&$70.4$&$47.9$&--&$16.3$&$46.4$\\
$\pi_{0.5}$~\cite{physicalintelligence2025pi05}&98.8&98.2&98.0&92.4&96.9&81.6&--&46.0&70.7\\
GR00T N1.6~\cite{nvidia2025groot16}&$97.7$&$98.5$&$97.5$&$94.4$&$97.0$&--&--&--&--\\
FabriVLA~\cite{yang2026fabrivla}&--&--&--&--&--&90.0&--&--&--\\
SUREFlow~\cite{islam2026sureflow}&94.8&91.0&93.8&90.2&92.5&88.3&--&--&--\\
LA4VLA-1B~\cite{lin2026la4vla}&93.4&98.2&98.0&93.4&95.8&87.5&--&--&--\\
\midrule
LIFD (Joint)&$82.8\pm2.4$&$81.7\pm2.7$&$80.2\pm1.9$&$77.4\pm2.2$&80.5&$69.4\pm2.8$&$0.20\pm0.06$&8.8&51.6\\
\textbf{LIFD (Staged)}&$93.6\pm1.7$&$92.8\pm1.5$&$91.3\pm2.1$&$88.6\pm1.8$&91.6&$79.8\pm1.9$&$0.14\pm0.03$&19.4&63.2\\
\bottomrule
\end{tabular*}
\end{table*}

\subsection{Policy Adaptation from Scene Tokens}
\label{sec:policy}
The Focus module applies slot attention~\cite{locatello2020object} to $\widehat{\mathbf S}_t$, producing $K_s$ compact slot features $\mathbf Z_t\in\mathbb R^{K_s\times d}$.
Slots compete for scene-token evidence and iteratively aggregate it.
Appendix~\ref{app:implementation} specifies scene-conditioned initialization and slot updates.
Slots have no prescribed one-to-one correspondence with physical objects.
The Do visuomotor policy predicts an action from slots, proprioception, and language:
\begin{equation}
\widehat{\mathbf A}_t=\pi_\psi
 \big([\operatorname{vec}(\mathbf Z_t);\mathbf P_t;\mathbf e_L]\big)
\label{eq:policy}
\end{equation}
During Staged policy learning, perception and completion remain frozen, while slot attention and the policy head learn from demonstrations:
\begin{equation}
\mathcal L_{\rm policy}=\mathbb E_t
 \|\widehat{\mathbf A}_t-\mathbf A_t^\star\|_1
 +\lambda_{\rm tv}\mathbb E_t
 \|\widehat{\mathbf A}_t-\widehat{\mathbf A}_{t-1}\|_2^2
\label{eq:policy_loss}
\end{equation}
Here $\mathbf A_t^\star$ is the demonstration action and $\lambda_{\rm tv}\geq0$ weights temporal smoothness.

\textbf{Joint versus Staged training.}
GeoVGGT-R remains frozen in both schedules.
Staged first learns the multi-view teacher, then trains recurrent-memory-conditioned RF completion toward the frozen teacher's scene-token targets. It finally trains slot attention and the policy with perception and completion frozen.
Joint simultaneously optimizes aggregation, RF completion and recurrent memory, slot attention, and the policy under a combined representation, completion, and control objective.
The comparison tests representation-first training against simultaneous optimization.

\section{Experiments}
\label{sec:experiments}

\textbf{Evaluation setting.}
\label{sec:setup}
We evaluate on LIBERO-Spatial, Object, Goal, and Long (LIBERO-10)~\cite{liu2023libero} and 50 MetaWorld tasks~\cite{yu2020meta}, with 50 demonstrations/task.
LIFD learns scene representations with multi-view supervision. Deployment requires one $224\times224$ RGB observation per step and recurrent memory.
LIBERO success follows the benchmark task criteria.
LIFD's LIBERO/MetaWorld results report mean $\pm$ SE over three training/evaluation seeds $\times$ 50 rollouts/task/seed (150/task); other methods report means only.
LIBERO Avg averages the four suite means.
Beyond policy demonstrations, GeoVGGT-R uses 100,000 simulated frames from RLBench, ManiSkill3, and RoboTwin and 10,000 real frames with geometric supervision.
Table~\ref{tab:main_results} summarizes benchmark results with method-specific sensing and training settings; Appendix~\ref{app:protocol} records these settings and result sources.

\newpage
\subsection{Control from Scene Memory}
\label{sec:sim_results}
LIFD (Staged) achieves 91.6\% LIBERO average success and 79.8\% MetaWorld success (Table~\ref{tab:main_results}).
On MetaWorld, LIFD (Staged) is 1.8 points below $\pi_{0.5}$ (81.6\%).
Staged training raises success over Joint by 10.8--11.2 points across LIBERO suites, 11.1 on average, and 10.4 on MetaWorld.

RoboTwin policy training uses 50 clean demonstrations/task, followed by 100 test rollouts/task/condition across 50 tasks~\cite{chen2025robotwin2}.
On Hard (Clean$\rightarrow$Random) and Easy (Clean$\rightarrow$Clean), Staged achieves 19.4\% and 63.2\%, compared with Joint's 8.8\% and 51.6\%.
Staged gains 10.6 points on Hard and 11.6 on Easy.
The 43.8-point decrease from Easy to Hard shows sensitivity to environment randomization after clean-demonstration training.

\subsection{Completion and Observation History}
\label{sec:ablations}
\begin{figure*}[t]
\centering
\includegraphics[width=0.95\textwidth]{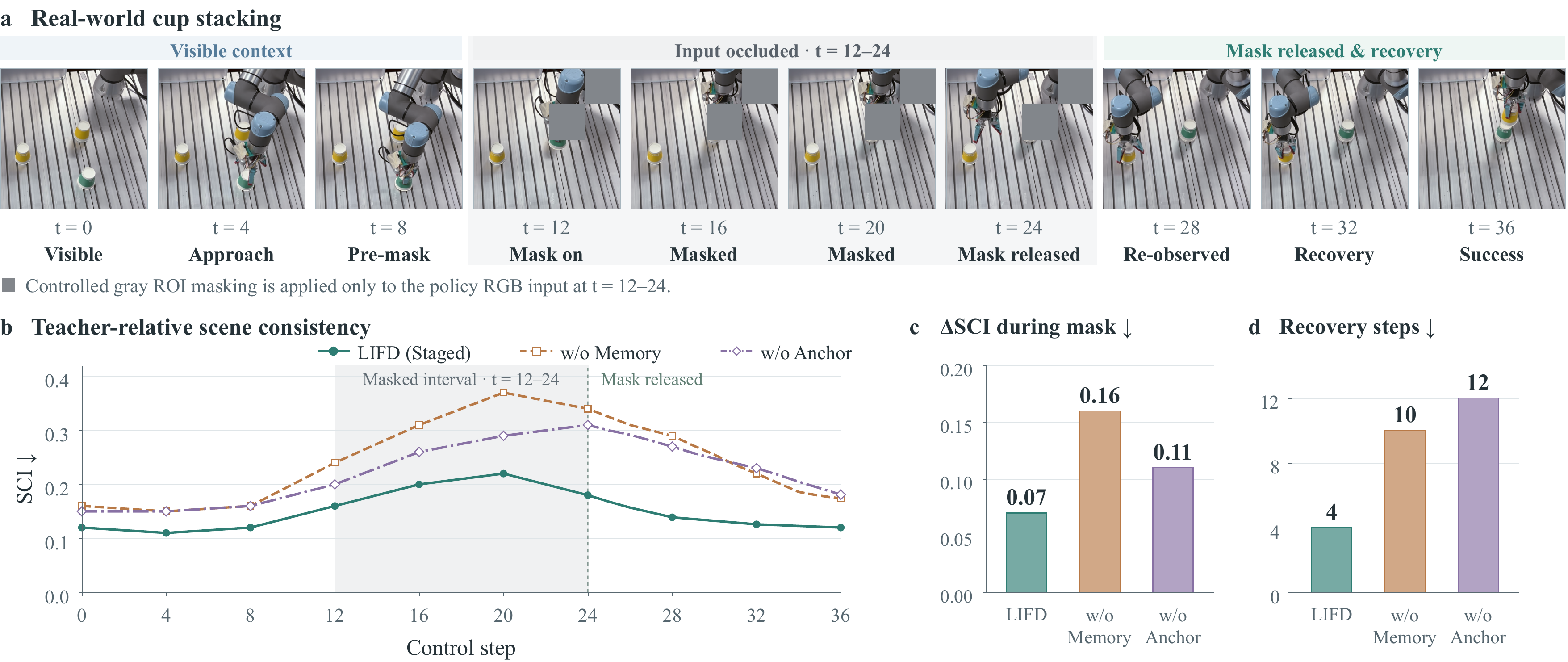}
\caption{\textbf{Real-world controlled occlusion.} A fixed gray ROI masks only the policy RGB input at steps 12--24 and is released immediately after step 24. (a) Keyframes through recovery and success. (b) Stepwise teacher-relative SCI. (c) Contradiction increase during masking. (d) Recovery steps after release. Summary definitions are in Appendix~\ref{app:protocol}.}
\label{fig:cup_intervention}
\end{figure*}

\textbf{Scene consistency.}
\label{sec:sci}
The Scene Contradiction Index (SCI, lower is better) is a teacher-relative latent diagnostic based on cosine agreement between Hungarian-matched generated and frozen multi-view teacher tokens.
Task success remains the primary evaluation metric.
Table~\ref{tab:main_results} reports LIBERO-Spatial SCI as mean $\pm$ SE over three training seeds (aggregation: Appendix~\ref{app:protocol}); baselines without compatible tokens have no entry.
Staged training reduces SCI from 0.20 to 0.14 alongside the increase in task success.

\textbf{Completion and anchoring.}
Table~\ref{tab:ablation} tests LIBERO-Spatial components against the 93.6\% full-model result in Table~\ref{tab:main_results}.
FiLM-only (no RF) retains current-view and memory FiLM fusion but removes generative completion, passing the deterministic representation directly to scene readout and control.
Success falls by 62.8 points and SCI rises from 0.14 to 0.62.
Removing the anchor eliminates direct current-view conditioning while retaining memory-conditioned AGCA ($-23.7$ points); removing memory retains only current-view conditioning ($-18.8$ points).
The Half-blind teacher uses the deployed policy's same-view evidence instead of richer multi-view evidence ($-42.5$ points).
Removing slot attention replaces iterative object-centric grouping with global scene readout ($-23.3$ points).
Disabling CFG preserves the full model and training-time condition dropout but integrates only the conditional vector field at inference; the 28.4-point decrease supports guidance for reliable completion.
\begin{table}[t]
\caption{LIBERO-Spatial ablations. SR: mean $\pm$ SE across three seeds; SCI: point estimates. Success-rate drops are relative to the 93.6\% full-model control in Table~\ref{tab:main_results}.}
\label{tab:ablation}
\centering
\small
\setlength{\tabcolsep}{5pt}
\begin{tabular}{lcc}
\toprule
Variant&SR (\%)$\uparrow$&SCI$\downarrow$\\
\midrule
Without anchor in AGCA&$69.9\pm3.9$&0.46\\
FiLM-only (no RF)&$30.8\pm5.7$&0.62\\
Half-blind teacher&$51.1\pm5.6$&0.24\\
Without memory in AGCA&$74.8\pm2.4$&0.19\\
Without slot attention&$70.3\pm1.3$&0.20\\
Without CFG&$65.2\pm4.2$&0.32\\
\midrule
\textbf{LIFD (Staged)}&$\mathbf{93.6\pm1.7}$&\textbf{0.14}\\
\bottomrule
\end{tabular}
\end{table}

\begin{table}[t]
\caption{UR5e success rates (\%, $\uparrow$): ten demonstrations and 50 trials per task family. Mean is the unweighted average across the four families; $\dagger$ follows Table~\ref{tab:main_results}.}
\label{tab:real_results}
\centering
\small
\setlength{\tabcolsep}{4pt}
\begin{tabular}{lccccc}
\toprule
Method&OM&SA&HTR&LR&Mean\\
\midrule
BC-R3M&36&12&32&4&21.0\\
OpenVLA-7B&54&32&64&12&40.5\\
Diffusion Policy&42&20&46&16&31.0\\
Lift3D$^\dagger$&48&34&48&8&34.5\\
\textbf{LIFD (Staged)}&\textbf{68}&\textbf{62}&\textbf{66}&\textbf{28}&\textbf{56.0}\\
\bottomrule
\end{tabular}
\end{table}

\begin{figure}[!b]
\centering
\includegraphics[width=\columnwidth]{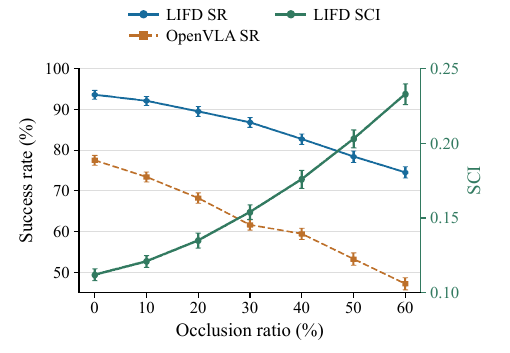}
\caption{\textbf{Occlusion on LIBERO-Spatial.} Left axis: success rate for LIFD (Staged) and OpenVLA-7B under identical random square masks. Right axis: LIFD's teacher-relative SCI.}
\label{fig:occlusion_curve}
\end{figure}

{\clubpenalty=10000
\textbf{Occlusion and memory interventions.}
\looseness=-1
In LIBERO-Spatial, random square occluders cover 0--60\% of each image in 10\% increments.
As occlusion increases, both success rates decline and LIFD's SCI rises (Fig.~\ref{fig:occlusion_curve}).
Zeroing recurrent memory at steps 10, 20, or 40 on LIBERO-Long reduces success with weights and noise schedules fixed, indicating a contribution from observation history.
\par}

{\clubpenalty=10000
We apply a fixed gray ROI mask only to the policy RGB input during cup stacking (Fig.~\ref{fig:cup_intervention}) at steps 12--24.
The mask is released immediately after step 24; the first displayed re-observation is step 28.
Full LIFD has the smallest masked-period contradiction increase ($\Delta\mathrm{SCI}_{\rm mask}$: 0.07 versus 0.16 without memory and 0.11 without anchor) and fastest recovery after release (four versus ten and twelve control steps).
Memory supports scene retention during masking, whereas current-view anchoring supports correction after re-observation.
\par}

\subsection{Physical Adaptation and Deployment Costs}
\label{sec:real}
We evaluate physical manipulation on a UR5e with RGB observations from a laterally mounted Intel RealSense D435i.
The evaluation spans twelve instances in four task families (Fig.~\ref{fig:tasks}).
OM requires retrieving a partly visible target, whereas HTR requires first uncovering it.
SA tests an ordered insertion and removal sequence, and LR tests relocation under viewpoint changes.
We adapt only LIFD's policy head from ten human demonstrations per family, keeping perception and completion frozen after representation learning with real images.
Each method is evaluated in 50 independent trials per family spanning predefined variations in poses, distractors, lighting, and camera extrinsics (Appendix~\ref{app:protocol}).

LIFD achieves 56.0\% macro-average success versus OpenVLA-7B's 40.5\%, a 15.5-point gain (Table~\ref{tab:real_results}).
The gain is largest on SA at 30 points and narrows to 2 points on HTR (one additional success in 50 trials).
LR remains difficult: LIFD succeeds in 14/50 trials (28\%) versus 8/50 (16\%) for Diffusion Policy, that task's strongest baseline.

\textbf{Inference cost.}
\label{sec:efficiency}
We use 50 RF integration steps by default.
On a single RTX~4090, reducing integration to 35 and 20 steps lowers end-to-end latency from 190\,ms to 130\,ms and 75\,ms, respectively.
These latencies correspond to 5.3, 7.7, and 13.3 policy updates per second.
Appendix~\ref{app:protocol} reports the full timing sweep (Table~\ref{tab:efficiency}).

\section{Discussion and Limitations}
\label{sec:limitations}
LIFD requires multi-view supervision during representation learning and iterative completion at inference.
Recovery of objects never observed within an episode remains untested.
Differences in pretraining resources constrain cross-method comparisons, and physical evaluation covers one robot platform and four task families.
Long-horizon relocation remains difficult.

\par\allowbreak
\begin{samepage}
\section{Conclusion}
\label{sec:conclusion}
We presented LIFD, which connects RGB-based geometric perception to visuomotor control through persistent, 3D-aware scene memory.
Multi-view supervision trains the scene-token representation; anchored RF completes it from current observations and recurrent memory, and slot features condition the policy.
Simulation and UR5e results support this scene representation for manipulation under partial observability.
\par
\end{samepage}

\useRomanappendicesfalse
\appendices
\section{Implementation Details}
\label{app:implementation}
\subsection{Geometric Supervision and Multi-View Aggregation}
\begin{figure*}[t]
\centering
\includegraphics[width=0.85\textwidth]{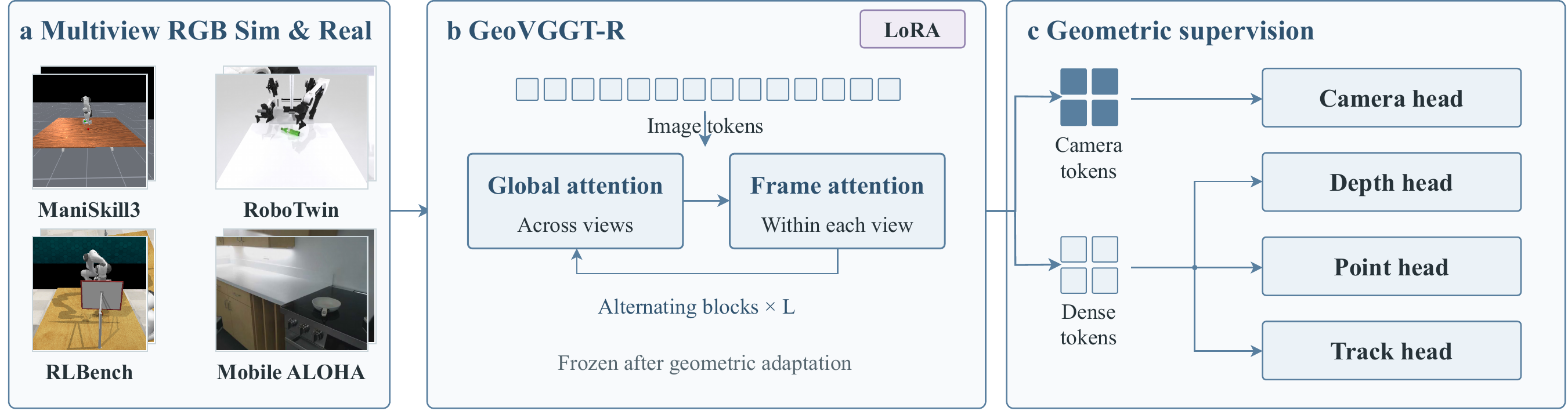}\par
\smallskip
\includegraphics[width=0.85\textwidth]{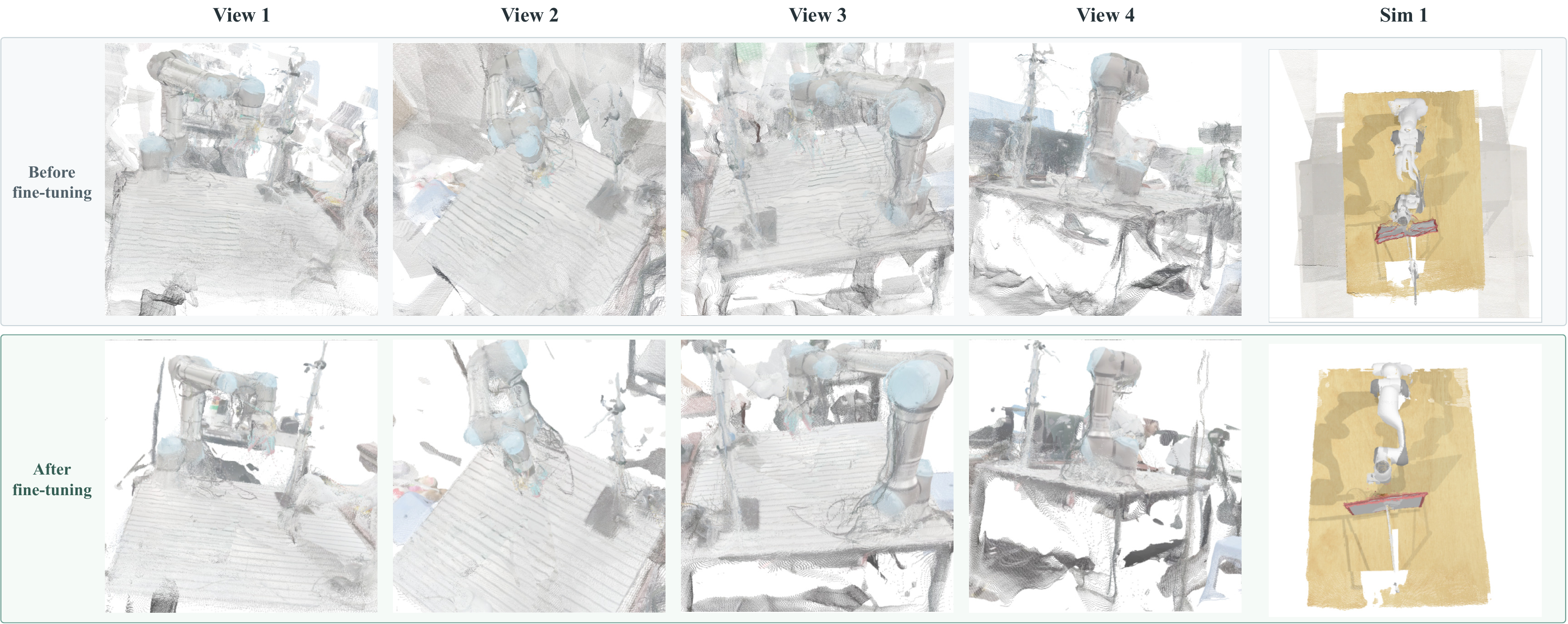}
\caption{\textbf{GeoVGGT-R.} Top: geometric supervision of camera and dense tokens. Bottom: qualitative reconstructions before and after adaptation from four real views and one simulated scene. In these examples, adaptation sharpens the robot-arm contours and reduces floating fragments.}
\label{fig:encoder_overcomplete}
\label{fig:recon}
\end{figure*}

Fig.~\ref{fig:encoder_overcomplete} shows supervision and reconstructions before and after geometric adaptation. The encoder minimizes
\begin{equation}
\mathcal L_{\rm enc}=\mathcal L_{\rm cam}+\mathcal L_{\rm depth}
 +\mathcal L_{\rm pmap}+\lambda_{\rm track}\mathcal L_{\rm track}
\label{eq:encoder}
\end{equation}
For view $v$, camera parameters $\mathbf g_v$ and track locations $\widehat{\mathbf y}_{j,v}$ are supervised by
\begin{equation}
\begin{gathered}
\mathcal L_{\rm cam}=\sum\nolimits_v\rho_\delta(\mathbf g_v-\mathbf g_v^{\rm gt}),\\
\mathcal L_{\rm track}=\sum\nolimits_{j,v}
 \|\widehat{\mathbf y}_{j,v}-\mathbf y_{j,v}^{\rm gt}\|_2,
\end{gathered}
\label{eq:camera_track_losses}
\end{equation}
The Huber loss is $\rho_\delta(\mathbf r)=\|\mathbf r\|_2^2/(2\delta)$ when $\|\mathbf r\|_2\leq\delta$, and $\|\mathbf r\|_2-\delta/2$ otherwise.
Depth and point-map heads are each trained with
\begin{equation}
\begin{aligned}
&\mathcal L_{\rm dense}=\sum_v\Big[
 \|\boldsymbol\sigma_v\odot(\widehat{\mathbf Y}_v-\mathbf Y_v^{\rm gt})\|_1\\
&\quad +\|\boldsymbol\sigma_v\odot(\nabla\widehat{\mathbf Y}_v-
 \nabla\mathbf Y_v^{\rm gt})\|_1-\lambda_{\rm conf}\sum_p\log\sigma_{v,p}\Big]
\end{aligned}
\label{eq:dense_geometric_loss}
\end{equation}
Here $\mathbf Y$ is depth or point-map data, $\nabla$ the spatial gradient, and $p$ a pixel index. Each head predicts separate softplus weights $\boldsymbol\sigma_v>0$, regularized by $\lambda_{\rm conf}>0$.

For support views $\mathcal I_{\rm supp}=\mathcal I\setminus\{I_a\}$, curriculum level $k$ admits
\begin{equation}
\begin{gathered}
\mathcal C^{(k)}={}\{\{I_a\}\cup U:\ U\subseteq\mathcal I_{\rm supp},\ |U|=k-1\},\\
 k=2,\ldots,n
\end{gathered}
\label{eq:view_curriculum}
\end{equation}
For encoded subset $\mathbf F_C$ and learned queries $\mathbf U^{(0)}=\mathbf Q_{\rm agg}\in\mathbb R^{M\times d}$, aggregation alternates cross- and self-attention:
\begin{gather}
\widetilde{\mathbf U}^{(\ell)}=\mathbf U^{(\ell-1)}+
 \operatorname{MHA}\!\bigl(\operatorname{LN}(\mathbf U^{(\ell-1)}),K(\mathbf F_C),V(\mathbf F_C)\bigr),\notag\\
\mathbf U^{(\ell)}=\widetilde{\mathbf U}^{(\ell)}+
 \operatorname{MHA}\!\bigl(\operatorname{LN}(\widetilde{\mathbf U}^{(\ell)}),
 K(\widetilde{\mathbf U}^{(\ell)}),V(\widetilde{\mathbf U}^{(\ell)})\bigr)
\label{eq:aggregation_blocks}
\end{gather}
$K,V$ are learned projections, LN is layer normalization, and $\mathbf S(C)=\mathbf U^{(L_{\rm agg})}$.

\subsection{Flow Conditioning and Slot Readout}
The time-conditioning MLP $f_{\rm time}$ modulates transformer normalization:
\begin{equation}
\begin{gathered}
\relax [\boldsymbol\gamma(u),\mathbf b(u)]=f_{\rm time}(u),\\
\operatorname{AdaLN}_u(\mathbf H)=
 \boldsymbol\gamma(u)\odot\operatorname{LN}(\mathbf H)+\mathbf b(u)
\end{gathered}
\label{eq:time_modulation}
\end{equation}
Scale and shift are broadcast over tokens; each block ends with pointwise feedforward layers.
Conditioning is dropped during training with probability $p_{\rm drop}$. At the same token state and flow progress, classifier-free guidance combines conditional and unconditional vector fields:
\begin{equation}
v^{\rm cfg}=v_{\rm cond}+w(v_{\rm cond}-v_{\rm uncond}),
\label{eq:cfg}
\end{equation}
$w$ is the guidance weight. Inference integrates $d\mathbf X/du=v^{\rm cfg}$ from noise ($u=0$) to scene tokens ($u=1$).

For $\mathbf x_i=\widehat{\mathbf s}_{t,i}$ and $\mathbf c_t=M^{-1}\sum_i\mathbf x_i$, initialize scene-conditioned Gaussian slots:
\begin{equation}
\begin{gathered}
\relax [\boldsymbol\mu_0,\log\boldsymbol\sigma_0]=g_{\rm init}(\mathbf c_t),\\
\mathbf z_k^{(0)}\sim\mathcal N\!\left(\boldsymbol\mu_0,
 \operatorname{diag}(\boldsymbol\sigma_0^2)\right),\quad k=1,\ldots,K_s
\end{gathered}
\label{eq:slot_init}
\end{equation}
Keys $\mathbf k_i=W_k\mathbf x_i$, values $\mathbf v_i=W_v\mathbf x_i$, and queries $\mathbf q_k^{(\ell)}=W_q\operatorname{LN}(\mathbf z_k^{(\ell)})$ define attention weights and aggregated token features:
\begin{equation}
\begin{gathered}
e_{ik}^{(\ell)}=(\mathbf q_k^{(\ell)})^\top\mathbf k_i/\sqrt d,\\
a_{ik}^{(\ell)}=\frac{\exp(e_{ik}^{(\ell)})}
 {\sum_{j=1}^{K_s}\exp(e_{ij}^{(\ell)})},\\
\mathbf r_k^{(\ell)}=\frac{\sum_{i=1}^{M}a_{ik}^{(\ell)}\mathbf v_i}
 {\sum_{i=1}^{M}a_{ik}^{(\ell)}+\varepsilon_{\rm attn}}
\end{gathered}
\label{eq:slot_evidence}
\end{equation}
Softmax normalizes attention across slots; each slot aggregates token values with stabilizer $\varepsilon_{\rm attn}>0$. The slot update is
\begin{equation}
\mathbf z_k^{(\ell+1)}=
 G_{\rm slot}(\mathbf z_k^{(\ell)},\mathbf r_k^{(\ell)})
 +f_{\rm res}(\operatorname{LN}(\mathbf z_k^{(\ell)})),
\label{eq:slot_update}
\end{equation}
$G_{\rm slot}$ is a GRU that updates each slot from its previous state and aggregated token features, and $f_{\rm res}$ is an MLP. Stacking the slots after $L_s$ iterations yields $\mathbf Z_t$.

\section{Evaluation Protocol and Diagnostics}
\label{app:protocol}
\begin{table}[H]
\caption{RTX~4090 inference cost. The policy update rate is the reciprocal of end-to-end latency, distinct from the robot command-interface rate.}
\label{tab:efficiency}
\centering
\small
\begin{tabular}{ccc}
\toprule
RF steps&Latency (ms)$\downarrow$&Updates/s$\uparrow$\\
\midrule
50&190&5.3\\
35&130&7.7\\
20&75&13.3\\
\bottomrule
\end{tabular}
\end{table}

\textbf{Evaluation counts.}
Our LIBERO/MetaWorld results use three seeds $\times$ 50 rollouts/task/seed (150/task).
LIBERO Avg averages four suites; MetaWorld Avg uses the benchmark's reported aggregate over difficulty tiers.
RoboTwin aggregates 100 rollouts/task/condition over 50 tasks.

\textbf{Training observations and baselines.}
Representation training uses seven views: front (anchor), top, back, left, right, front-left, and front-right.
Teacher aggregation uses progressively larger anchored subsets and EMA coefficient $[0.99,0.999]$.

Baselines include pretrained 2D features~\cite{nair2022r3m,majumdar2023we}, action diffusion~\cite{chi2025diffusion}, VLAs~\cite{kim2024openvla,black2410pi0}, and adapted 3D policies~\cite{shridhar2023perceiver,ze20243d,jia2025lift3d}.
RoboTwin entries retain the sensing and clean-data training settings reported by the official leaderboard~\cite{robotwin2026leaderboard}.
\textbf{Result sources.}~LIBERO ACT uses STAR's 50-demonstration/task results~\cite{li2025star}; RDT-1B uses LatBot~\cite{li2025latbot}.
MetaWorld reference results use Evo-SOTA standard-evaluation averages~\cite{evosota2026metaworld}, including LA4VLA-1B MixPT~\cite{lin2026la4vla}.
OpenVLA-OFT uses external and wrist cameras and filtered demonstrations~\cite{kim2025oft}; LIBERO results for $\pi_{0.5}$ and GR00T N1.6 follow the baseline results reported in ABot-M0~\cite{yang2026abotm0}.
RoboTwin's board includes Single-task SFT and Co-train tracks~\cite{robotwin2026leaderboard}: $\pi_{0.5}$ trains across 50 clean-data tasks, while the other listed public baselines use per-task fine-tuning.

\textbf{Physical baseline adaptation.}
BC-R3M freezes its pretrained encoder and trains only the policy head.
Lift3D also adapts only its policy head, with all other components frozen.
OpenVLA-7B and Diffusion Policy are fine-tuned on the same ten demonstrations per family.

\textbf{Scene Contradiction Index.}
The same frozen multi-view teacher, learned from the training data, provides $\mathbf S_t^\star$ for offline SCI evaluation across variants.
Hungarian permutation $\Pi_t$ minimizes generated--teacher cosine distance. With episode length $H$ and warm-up $W=10$,
\begin{equation}
\begin{gathered}
\mathrm{SCI}=1-\frac{1}{H-W}\sum_{t=W+1}^{H}c_t,\\
c_t=\frac{1}{M}\sum_{m=1}^{M}
 \cos\!\left(\widehat{\mathbf s}_{t,\Pi_t(m)},\mathbf s_{t,m}^\star\right)
\end{gathered}
\label{eq:sci}
\end{equation}
Table~\ref{tab:main_results} reports LIBERO-Spatial SCI as mean $\pm$ SE over three training seeds.
Within each seed, rollout SCI is averaged within each task and then equally across tasks.
Table~\ref{tab:ablation} reports SCI point estimates.

\textbf{Cup-intervention summaries.}
Fixed diagnostic cameras record each cup-stacking rollout in sync with the main D435i.
At each control step, the frozen teacher aggregates synchronized, unmasked multi-view observations into $\mathbf S_t^\star$ for offline SCI.
Only the policy's main-camera RGB input is masked; diagnostic views do not enter online inference or control.
Fig.~\ref{fig:cup_intervention} uses stepwise $\mathrm{SCI}_t=1-c_t$ before rollout averaging.
Let $\mathrm{SCI}_{\rm pre}$ be the mean over steps $\{0,4,8\}$.
$\Delta\mathrm{SCI}_{\rm mask}$ is the mean at $\{12,16,20,24\}$ minus $\mathrm{SCI}_{\rm pre}$.
Recovery is the first post-release control step with $\mathrm{SCI}_t\leq1.2\,\mathrm{SCI}_{\rm pre}$, with $T_{\rm rec}=t_{\rm recover}-24$.
Recovery uses control-step resolution; displayed keyframes are four steps apart.

\textbf{Physical tasks and hardware.}
OM extracts a pink block partly hidden by a ceramic bowl wall, varying occlusion depth, bowl placement, and distractors.
SA inserts green then pink blocks into a container and removes them in prescribed order, varying order, container depth, and initial poses.
HTR moves a plush toy fully occluding a green target before grasping the target, varying occluder type, mass, and target placement.
LR stacks a green block on a red block under camera-extrinsic shifts of 2--10\,cm/$5$--$20^\circ$.
Success requires target extraction (OM), the correct insertion/removal sequence (SA), occluder movement followed by hidden-target retrieval (HTR), or completion of the relocation/stacking goal (LR).

The UR5e uses a compliant three-finger gripper.
The laterally mounted Intel RealSense D435i faces the workspace center and captures RGB at 30\,Hz. Depth is used only for visualization and annotation.
Cartesian position control sends 6-DoF end-effector poses through Real-Time Data Exchange (RTDE) at 125\,Hz, independently of policy inference throughput.

\bibliographystyle{IEEEtran}
\bibliography{main}
\end{document}